\documentclass[10pt,twocolumn,letterpaper]{article}

\usepackage{iccv}
\usepackage{times}
\usepackage{epsfig}
\usepackage{graphicx}
\usepackage{amsmath}
\usepackage{amssymb}

\usepackage{booktabs}
\usepackage{microtype}
\usepackage{wrapfig}
\usepackage{color}
\usepackage{soul}
\usepackage{xcolor}

\def\figurePath{images/}
\def\myfigure#1#2{\begin{figure}[t]\centering\includegraphics*[width = \linewidth]{\figurePath#1}\caption{#2}\label{fig:#1}\end{figure}}
\def\mycfigure#1#2{\begin{figure*}[t]\centering\includegraphics*[clip, width = \linewidth]{\figurePath#1}\caption{#2}\label{fig:#1}\end{figure*}}

\def\mysection#1#2{\section{#1}\label{sec:#2}}

\def\mysubsection#1#2{\subsection{#1}\label{sec:#2}}

\newcommand{\refSec}[1]{Sec.~\ref{sec:#1}}
\newcommand{\refFig}[1]{Fig.~\ref{fig:#1}}

\newcommand{\refTbl}[1]{Table~\ref{tbl:#1}}

\definecolor{unsurecolor}{rgb}{1,.85,.7}
\definecolor{changedcolor}{rgb}{.85,1,.7}

\soulregister\ref7
\soulregister\cite7
\soulregister\citeetal7
\soulregister\etal0
\soulregister\eg0
\soulregister\scene1
\soulregister\refTbl1

\DeclareGraphicsExtensions{.pdf,.ai,.psd,.jpg}
\newcommand{\myparagraph}[1]{
\noindent
\textbf{#1}
\hspace{0.1cm}
}

\usepackage[breaklinks=true,bookmarks=false]{hyperref}
\iccvfinalcopy 

\def\iccvPaperID{2246} 

\begin{document}

\title{What Is Around The Camera?}

\author{Stamatios Georgoulis\\
KU Leuven\\
\and
Konstantinos Rematas\\
University of Washington\\
\and
Tobias Ritschel\\
University College London\\
\and 
Mario Fritz\\
MPI for Informatics\\
\and 
Tinne Tuytelaars\\
KU Leuven\\
\and
Luc Van Gool\\
KU Leuven \& ETH Zurich
}

\maketitle

\begin{abstract}
How much does a single image reveal about the environment it was taken in?
In this paper, we investigate how much of that information can be retrieved from a foreground object, combined with the background (\ie the visible part of the environment). 
Assuming it is not perfectly diffuse, the foreground object acts as a complexly shaped and far-from-perfect mirror. 
An additional challenge is that its appearance confounds the light coming from the environment with the unknown materials it is made of. 
We propose a learning-based approach to predict the environment from multiple reflectance maps that are computed from approximate surface normals. 
The proposed method allows us to jointly model the statistics of environments and material properties. 
We train our system from synthesized training data, but demonstrate its applicability to real-world data. 
Interestingly, our analysis shows that the information obtained from objects made out of multiple materials often is complementary and leads to better performance.

\end{abstract}

\mysection{Introduction}{Introduction}

Images are ubiquitous on the web. 
Users of social media quite liberally contribute to this fast growing pool, but also companies capture large amounts of imagery (e.g. street views). 

Simultaneously, users have grown aware of the risks involved, forcing companies to obfuscate certain aspects of images, like faces and license plates.

\myfigure{Teaser}{
Given approximate surface normals, our approach predicts a truthful reconstruction of the environment from a single low dynamic range image of a (potentially multi-material) object.
}

We argue that such images contain more information than people may expect. 
The reflection in a foreground object combined with the background - the part of the environment revealed directly by the image - can be quite telling about the entire environment, i.e. also the part behind the photographer. 
Apart from privacy issues, several other applications can be based on the extraction of such environment maps. 
Examples include the relighting of the foreground object in a modified environment, the elimination of the photographer in the reflection by the foreground object, or figuring out that a photo might have been nicely framed, but that the actual surroundings may not be quite as attractive as one might have been lead to believe (\eg browsing through real estate). 

By combining large amount of synthesized data and deep learning techniques, we present an approach that utilizes partial and ambiguous data encoded in surface reflectance in order to reconstruct a truthful appearance of the surrounding environment the image was taken in.
While we assume a rough scene geometry to be given, we do not make any strong assumptions on the involved materials or the surrounding environment.
Yet, we obtain a high level of detail in terms of color and structure of that environment. 
Beyond the new level of reconstruction quality obtained by our method, we show that our learned representations and final predictions lend to a range of application scenarios. 
Experiments are conducted on both synthetic and real images from the web.

Prior work on estimating the illuminating environment has either focused on retrieving a template map from a database (e.g.~\cite{Karsch2014}) or recovering key elements to approximate it (e.g.~\cite{Lalonde2017}), but both directions are limited 
in terms of the achievable fidelity.
Equally, prior work on inverse rendering \cite{mahajan2008theory, Lombardi2012, Lombardi2015} has the potential to derive arbitrary maps with radiometric accuracy, but to date are not able to do so with accurate appearance. 
In this spirit, we seek a different tradeoff. 
By not constraining our solution to be radiometrically accurate, we investigate how far we can push the quality of overall appearance of the map. 
We hope that these two approaches can complement each other in the future.

Our work can be seen as an extension of the ideas presented in \cite{georgoulis2016delight} and is the first to recover such highly detailed appearance of the environment. 
We achieve this in a learning-based approach that enables joint learning of environment and material statistics. 
We propose a novel deep learning architecture that combines multiple, partially observed reflectance maps together with a partial background observation to perform a pixel-wise prediction of the environment.

\mysection{Previous work}{PreviousWork}

Object appearance is the result of an intriguing jigsaw puzzle of unknown shape, material reflectance, and illumination.
Decomposing it back into these intrinsic properties is far from trivial~\cite{Barrow1978}.
Typically, one or two of the intrinsic properties are assumed to be known and the remaining one is estimated.
In this work, we focus on separating materials and illumination when the partial reflectance maps of multiple materials seen under the same illumination plus a background image are known.
Such an input is very typical in images, yet not so often studied in the literature.

Key to this decomposition into intrinsic properties is to have a good understanding of their natural statistics.
Databases of material reflectance \cite{Dana1999, Matusik2003, Bell2014} and environmental illumination \cite{Debevec1998, Dror2001} allow the community to make some first attempts. 
Yet, exploiting them in practical de-compositions remains challenging.

\textbf{Beyond image boundaries}
Revealing "hidden" information about the environment in which an image was taken, has been used for dramatic purposes in movies and tv shows, but it has also attracted research interest. 
In~\cite{Torralba2014}, Torralba~\etal are using a room as a camera obscura to recover the scene outside the window. 
Nishino~\etal~\cite{Nishino2006} estimate panoramas from the environment reflected in a human eye. 
On the other hand, \cite{Zhang13} estimate a plausible panorama from a cropped image by looking at images with similar structure, while
\cite{Wang2014} extend the content of the image by high level graph matching with an image database. 
Moreover, a single image can contain hidden "metadata" information, such as GPS coordinates~\cite{Hays2008} or the photographer identity~\cite{Thomas2016}.

\textbf{Reflectance maps}
\emph{Reflectance maps} \cite{Horn1979} assign appearance to a surface orientation for a given scene/material, thus combining surface reflectance and illumination.
Reflectance maps can be extracted from image collections \cite{Haber2009}, from a known class \cite{Rematas2014, RematasPAMI2017}, or using a CNN \cite{Rematas2016}.
In computer graphics, reflectance maps are used to transfer and manipulate appearance of photo-realistic or artistic ``lit spheres'' \cite{Sloan2001} or ``MatCaps'' \cite{Zbrush}.
Khan~\cite{Khan2006} made diffuse objects in a photo appear specular or transparent using manipulations of the image background that require manual intervention.

\textbf{Factoring illumination}
Classic intrinsic images factor an image into shading and reflectance \cite{Barrow1978}.
Larger-scale acquisition of reflectance \cite{Matusik2003} and illumination \cite{Debevec1998} have allowed to compute their statistics \cite{Dror2001} helping to better solve inverse and synthesis problems.
Nevertheless, intrinsic images typically assume diffuse reflectance.
Barron and Malik~\cite{Barron2015a} decompose shaded images into shape, reflectance and illumination, but only for scalar reflectance, \ie diffuse albedo, and for limited illumination frequencies.
Richter~\etal~\cite{Richter2015} estimated a diffuse reflectance map represented in spherical harmonics using approximate normals and refined the normal map using the reflectance map as a guide, but their approach is suitable to represent low-frequency illumination, while our environment maps reproduce fine details. 
 
Separating material reflectance (henceforth simply referred to as 'material') and illumination was studied by Lombardi and Nishino~\cite{Lombardi2012, Lombardi2015, LombardiN16}.
In \cite{Lombardi2012, Lombardi2015}, they present different optimization approaches that allow for high-quality radiometric estimation of material and illumination from known 3D shape and a single HDR RGB image, whereas in \cite{LombardiN16}, they use multiple HDR RGBZ images to acquire material and illumination and refine shape using a similar optimization-based framework also handling indirect lighting.
Here, we address a more general problem than these approaches: they consider one or more objects with a single, unknown material on their surface (homogeneous surface reflectance) observed under some unknown natural illumination.
However, most real-life objects are made of multiple materials and as noted in \cite{lensch2003image, zickler2006reflectance, lombardi2012single} multiple materials help to estimate surface reflectance under a single point light source.
Furthermore, they throw away the image background that is naturally captured in their images anyway.

In this paper, we assume that the objects consist of any number of materials (we investigate up to 5), that they are segmented into their different materials as well as from the background, and that the reflectance maps of all materials are extracted (conditions that are met by \eg using a Google Tango phone as in \cite{zhang2016emptying}). 
We then ask how these multiple materials under the same non-point light illumination plus the background can help a deep architecture to predict the environment.
Note that, unlike \cite{Lombardi2012, Lombardi2015} our primary goal is to estimate a "textured" environment map that is perceptually close to ground truth and can provide semantic information (\eg the object is placed in a forest); later in \refSec{Results} we examine if we can recover the dynamic range too as a secondary goal.
Most importantly, we aim to do this starting from standard LDR images, as the HDR ones used in \cite{Lombardi2012, Lombardi2015, LombardiN16} imply the capture of multiple exposures per image making the capturing process impractical for non-expert users.

Barron \etal \cite{Barron2015b} made use of similar data to resolve spatially-varying, local illumination.
While ours is spatially invariant (distant), we can extract it both with more details, in HDR and from non-diffuse surfaces.

Earlier work has also made use of cues that we did not consider, such as shadows~\cite{sato2003illumination}, as they may only be available in some scenes.
The work of \cite{lalonde2012estimating, Lalonde2017} has shown how to fit a parametric sky model from a 2D image, but cannot reproduce details such as buildings and trees and excludes non-sky, \ie indoor settings.
Karsch \etal \cite{Karsch2014} automatically inferred environment maps by selecting a mix of nearest neighbors (NN) from a database of environment maps that can best explain the image assuming diffuse reflectance and normals have been estimated.
They demonstrate diffuse relighting but specular materials, that reveal details of a reflection, hardly agree with the input image as seen in our results section. 


\mysection{Overview}{Overview}

\myfigure{Spheres_small}{
\textbf{a)} Illustration of \refSec{Overview}.
\textbf{b)} Converting 3 example input pixels into reflectance map pixels using normals and segmentation.
}

We formulate our problem as learning a mapping from $n_\mathrm{mat}$ partial LDR reflectance maps \cite{Horn1979} and a background LDR image to a single consensual HDR environment map.
In particular, we never extract illumination directly from images, but indirectly from reflectance maps.
We assume the reflectance maps were extracted using previous work \cite{Eigen2015, Wang2015, Li2015, Rematas2016}.
In our datasets, 
we analyze the limits of what reflectance map decomposition can do and we do not consider the error introduced by the reflectance map itself.

A reflectance map $L_\mathrm{o}(\omega)$ represents the appearance of an object of a homogeneous material under a specific illumination.
Under the assumptions of (i) a distant viewer, (ii) distant illumination, (iii) in the absence of inter-reflections or shadows (convex object) and (iv) a homogeneous material, the appearance depends only on the surface orientation $\omega$ in camera space and can be approximated as a convolution of illumination and material (\ie BRDF) \cite{ramamoorthi2001signal}.

The full set of orientations in $\mathbb R^3$ is called the 3D \emph{Gauss} sphere $\Omega$ (the full circle in \refFig{Spheres_small}~a and b).
Note, that only at most half of the orientations in $\mathbb R^3$ are visible in camera space, \ie the ones facing into the direction of the camera.
This defines the positive Gauss sphere $\Omega^+$ (the brown half-circle in \refFig{Spheres_small}~a).
Also note, that due to the laws of reflections, surfaces oriented towards the viewer also expose illumination coming from behind the camera.
The ideal case is a one-material spherical object, that completely contains all observable normals. 
When its surface behaves like a perfect mirror, that is even better. 
Then a direct (but partial) environment map is directly observable. 
In practice, we only observe some orientations for some materials and other orientations for other materials.
Sometimes, multiple materials are observed for one orientation, but it also happens that for some orientations, no material might be observed at all.
Moreover, the materials tend to come with a substantially diffuse component in their reflectance, thus smearing out information about the environment map. 
In \refFig{Spheres_small}~a, the brown part shows the half-sphere of the reflectance map and the yellow part within shows the object normals actually observed in the image, for the example object in the figure.

A second piece of input comes from the background.
The visible part of the background in the image shows another part of the illumination, this time from the negative half sphere.
In \refFig{Spheres_small}~a, the visible part of the background is shown in red, the rest - occluded by the foreground - in blue. 

The illumination $L_\mathrm{i}(\omega)$ we will infer from both these inputs covers the full sphere of orientations $\Omega$ (the full circle in \refFig{Spheres_small}~a).
Other than the reflectance map, it typically is defined in world space as it does not change when the viewer's pose changes. 
For the actual computations, both the input (partial reflectance maps and partial background) and the output (illumination) are represented as two-dimensional images using a latitude-longitude parameterization.

The mapping $f := L_\mathrm o\rightarrow L_\mathrm i$ we seek to find is represented using a deep CNN.
We propose a network that combines multiple convolutional stages - one for each reflectance map, that share weights, and another one for the background - with a joint de-convolutional stage that consolidates the information into a detailed estimate of the illumination.

The training data consists of tuples of reflectance maps $l_\mathrm o$ with a single background image, \ie inputs, and a corresponding illumination $l_\mathrm i$, \ie output.


\mysection{Dataset}{Dataset}

Our dataset consists of synthetic training and testing data (\refSec{TrainingData}) and a manually-acquired set of test images of real objects captured under real illumination (\refSec{TestData}).
Upon publication, the dataset and code will be made available.

\myfigure{DataSet_small}{
Example images from our dataset.
\textbf{1st col:} Synthetic images of cars with a single material.
\textbf{2nd col:} Synthetic images of cars with multiple materials.
\textbf{3rd col:} Photographs of spheres with a single material.
\textbf{4th col:} Photographs of toy cars with a single material.
\textbf{5th col:} Photographs of toy cars with multiple materials.
}

\mysubsection{Synthetic data}{TrainingData}
We now explain how to synthesize train and test data. 

\myparagraph{Rendering}
Images are rendered at a resolution of $512 \times 512$ using variations of geometry, illumination, materials and views.
The geometry is a random object from the ShapeNet~\cite{ChangFGHHLSSSSX15} class ``car''.
Later, we show results of our pipeline for both cars and other shapes though (\eg \refFig{Teaser}).
As large 3D shape datasets from the Internet do not come with a consistent segmentation into materials, we perform a simple image segmentation after rasterization.
To this end, we perform $k$-means clustering ($k=n_\mathrm{mat}$) based on positions and normals, both weighted equally and scaled to the range $(-1,1)$, to divide the shapes into three regions, to be covered with three different `materials'.
Per-pixel colors are computed using direct (no global illumination and shadows) image-based illumination \cite{Debevec1996}.
We also store per-pixel ground-truth positions and normals.
As materials we used the 100 BRDF samples from the MERL database \cite{Matusik2003}.
The illumination is randomly selected from a set of $105$ publicly available HDR environment maps that we have collected.
The views are sampled randomly over the sphere, with a fixed field-of-view of $30$ degrees.
Synthetic examples can be seen in the first two columns of \refFig{DataSet_small}.

\myparagraph{Extracting reflectance maps}
The pixel $j$ in the reflectance map of material $i$ is produced by averaging all pixels with material $i$ and orientation $\omega_j$.
This is shown for three pixels from three different materials in \refFig{Spheres_small}~b.
The final reflectance maps contain $128\times 128$ pixels. 
These are typically partial with sometimes as little as 10\% of all normals observed (see some examples in \refFig{Architecture}).
Even sparser inputs have not been studied (\eg hallucination works \cite{Georgoulis2015}). 

\myparagraph{Background extraction}
The background is easily identified for these synthetic cases, by detecting all pixels where the geometry did not project to.
To make the network aware of depth-of-field found in practice, the masked background is filtered with a 2D Gaussian smoothing kernel ($\sigma=2$).

\myparagraph{Building tuples}
To test our approach for an arbitrary number of materials, $n_\mathrm{mat}$, we build tuples by combining $n_\mathrm{mat}$ random reflectance maps extracted from images with a single material.
For each tuple we make sure to use mutually exclusive materials observed under the same illumination.

\myparagraph{Splitting}
For the single-material case, from the 60\,k synthetic images generated, 54\,k are used for training and 6\,k for testing.
Note that, no environment map is shared between the two sets - 94 for training and 11 for testing, with the split generated randomly once.
For the multi-material case, we used the same protocol (identical split) for two different sub-cases: multiple single-material objects (\ie the tuples), and single multi-material objects (\eg \refFig{DataSet_small}, 2nd column).
In both cases, the environment maps are augmented by rotations around the y axis, which is essential for preventing the networks to memorize the training samples.

\mysubsection{Real data}{TestData}
While training can be done on massive synthetic data, the network ultimately is to be tested on real images.
To this end, we took photographs of naturally illuminated single-material and multi-material objects with known geometry. 
All images in this set - 112 in total - were used for testing and never for training.
Moreover, all 3D models, materials and illuminations in this set are unknown to the train set.

\myparagraph{Capture}
The images are recorded in LDR with a common DSLR sensor at a resolution of 20\,M pixels and subsequently re-scaled to match the training data. 
To compare with reference, we also acquired the environment map in HDR using a chrome sphere.
\begin{wrapfigure}{c}{0.16\textwidth} 
\includegraphics[width=0.16\textwidth]{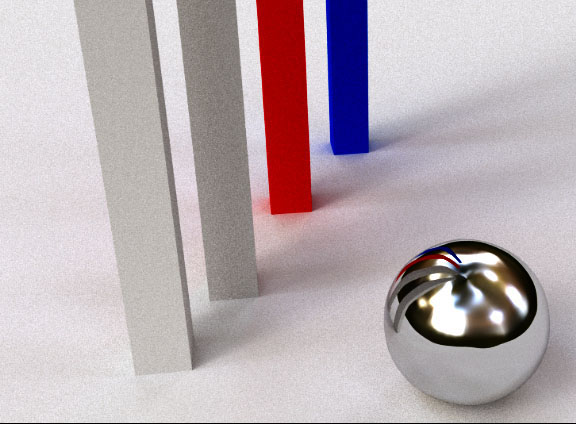}
\end{wrapfigure}
We recorded 7 $f$-stops \cite{debevec2008recovering}, which is a compromise between shadow fidelity and capture time, but as the inset figure shows - an example rendering with an estimated environment map (\refFig{RealResults_small}, row 2) - it is enough to produce clearly directed shadows.
Three variants were acquired: spheres, single-material objects and multi-material objects (see \refFig{DataSet_small}).
For the single-material case, 84 images were taken, showing 6 spheres and 6 toy cars with different materials each and placed under 7 different illuminations.
The multi-material data comprises of 30 images, showing 6 different objects (4 cars and 2 non-cars), each painted with 3 materials, captured under 9 different illuminations (6 and 3 respectively). 
Some materials repeat, as overall 12 different materials were used.

\myparagraph{Extracting reflectance maps and background}
From all images, reflectance maps are extracted in the same way as for the synthetic images.
Per-pixel normals are produced using virtual replica geometry from online repositories or scanned using a structured-light scanner. 
These models were manually aligned to the 2D images.
Material and background segmentation was also done manually for all images.


\mysection{Network Architecture}{DeMatNet} 

\mycfigure{Architecture}{
	Our CNN architecture \emph{(from left to right)}.
    First, the background image is encoded using one independent sub-network \emph{(blue)}.
    Next, each partial reflectance map is encoded using $n_\mathrm{mat}$ de-reflection sub-networks that share parameters\emph{(blue)}.
    Finally, these two sources of information are fused in a de-convolution network \emph{(orange)}.
    Here, information from all levels of the partial reflectance maps is included \emph{(violet)} as well as the global encoding of the background \emph{(green)}.
    Sub-network details are found in the text and supplementary material. 
}

Our network consists of three parts (\refFig{Architecture}) - some of them identical in structure and some sharing weights.
First, there is a convolutional \emph{background} network.
Second, $n_\mathrm{mat}$ convolutional \emph{de-reflection} networks that share parameters but run on the reflectance maps of different materials.
Third, a final de-convolutional \emph{fusion} network takes as input intermediate stages as well as end results from all reflectance nets, together with the result of the background net, to produce the HDR environment map as an output.
All parts are trained jointly end-to-end using an L1 loss on the illumination samples, after applying the natural logarithm and converting them to CIE Lab space.
We experimentally found that these choices nicely balance between learning the dynamic range, structure and color distribution of the environment map.

\textbf{Background network}
Input to the background network (blue part in \refFig{Architecture}, a) is an LDR background image in full resolution \ie 128$\times$128 converted to CIE Lab space.
The output is a single, spatially coarse encoding of resolution 4$\times$4.
The reduction in spatial resolution is performed as detailed in supplementary material. 
Only the final output of the encoding step will contribute to the fusion (\refFig{Architecture}, d).

\textbf{De-reflection network}
The de-reflection network (blue parts in \refFig{Architecture}, b) consumes partial, LDR environment maps also converted to CIE Lab space, where undefined pixels are set to black.
It has the same structure as the background network.
It starts with the full, initial reflectance map at a resolution of 128$\times$128 and reduces to a spatial resolution of 4$\times$4.
We can support an arbitrary number of materials $n_\mathrm{mat}$; however the network needs to be trained for a specific number of materials.
In any case, the de-reflection networks are trained with shared parameters (siamese architecture; locks in \refFig{Architecture}).
We want each of these networks to perform the same operations so the reflectance maps do not need to come in a particular order. 

\textbf{Fusion network}
The fusion network (\refFig{Architecture}, e) combines the information from the background and the de-reflection network.
The first source of information are the intermediate representations from the reflectance maps (violet, \refFig{Architecture}, c).
They are combined using plain averaging with equal weights.
This is done at each scale of the de-reflection, respectively, at each level of the fusion.
The second source of information is the background (green in \refFig{Architecture}, d).
Here, only a single, spatial level is considered, i.e. that of its output. 
This encoding is concatenated with the average of the encodings from all reflectance maps on the coarsest level (\ie their spatial resolution matches).
Result of this sub-network is the final 64$\times$64 HDR environment map.

\textbf{Missing pixels} 
Although the input reflectance maps are partial, the output environment map is complete.
This is an advantage of the proposed CNN architecture, as in the absence of information the fusion network learns how to perform sparse data interpolation for the missing pixels, as in \cite{Rematas2016}.
Our learning-based approach naturally embeds priors from the statistics of reflectance and illumination in the training data (\eg the sky is blue and always on top) which is hard to model with physics-based analytic formulas \cite{Lombardi2012, Lombardi2015}.

\textbf{Training details} We use mini-batch (size 4) gradient descent, a log-space learning rate of (-3, -5, 50), a weight decay of .0005, a momentum of .95 and train for 50 iterations.


\mysection{Results}{Results}

\mysubsection{Quantitative evaluation}{Quantitative}
We quantify to which extent our approach can infer the environment from an LDR input.
Since we are interested in estimating an environment map that is closer to the truthful illumination and does not only have HDR content, 
we use the perceptualized DSSIM \cite{wang2004image} (less is better) as our evaluation metric.
This metric captures the structural similarity between images \cite{Narihira2015b, Rematas2016, pathakArxiv15, chen2013simple}, that is of particular importance when the environment's reflection is visible in a specular surface, such as the ones we target in this paper.
The evaluation protocol includes the next steps:
The $.90$-percentile is used to find a reference exposure value for the ground truth HDR environment map.
We then apply the same tone-mapper with this authoritative exposure to all HDR alternatives.
This way we ensure that we achieve a fair comparison of the different alternatives also w.r.t. HDR.

\textbf{Model variants and baselines}
The results of different variants of our approach and selected baseline methods are presented in terms of performance in \refTbl{Results} and visual quality in \refFig{Alternatives_small}.
Below, we describe the different approaches:\\
$\bullet$ \textsc{Singlet} uses only a single reflectance map, \ie our de-reflection network with $n_\mathrm{mat}=1$, but without background.\\
$\bullet$ \textsc{Singlet+BG} also uses a single reflectance map, as before, but includes the background network too.\\
$\bullet$ {\color{gray} \textsc{Best-Of-Singlets}} executes the $n_\mathrm{mat}=1$ de-reflection-plus-background network for each singlet of a triplet individually and then chooses the result closest to the reference by an oracle (we mark all oracle methods in gray).\\
$\bullet$ {\color{gray} \textsc{Nearest Neighbor}} is equivalent - an upper bound - to \cite{Karsch2014}, but as their dataset of illuminations is not publicly available, we pick the nearest neighbor to ground-truth from our training dataset by an oracle so that DSSIM is minimized.\footnote{Note that, \cite{Karsch2014} is the only other published work with the same input (a single LDR image) and output (a HDR non-parametric environment map).}\\
$\bullet$ {\textsc{Mask-Aware Mean}} executes $n_\mathrm{mat}=1$ de-reflection-plus-background network for each singlet of a triplet individually and then averages the predicted environment maps based on the sparsity masks of the input reflectance maps.\\
$\bullet$ {\textsc{Triplet}} combines the information from three reflectance maps via our de-reflection ($n_\mathrm{mat}=3$) and fusion networks, without using background information.\\
$\bullet$ {\textsc{Triplet+BG}} represents our full model that combines the de-reflection ($n_\mathrm{mat}=3$), fusion and background networks.

\begin{table*}[t]
	\centering
	\caption{DSSIM error (less is better) for different variants \emph{(rows)} when applied to different subsets of our test set \emph{(columns)}.
    The best alternative is shown in \textbf{bold}.
    Oracle analysis using ground-truth information are shown in {\color{gray}gray}.
    Variant images are seen in \refFig{Alternatives_small}.
	}
    \setlength{\tabcolsep}{5.0pt}
	\begin{tabular}{rrrrrrr}
	&\multicolumn{2}{c}{$\;\;$--------- Synthetic ----------}
    &\multicolumn{4}{c}{------------------------------- Real -------------------------------} 
    \\
    & Cars (Single) & Cars (Multi) & Spheres & Cars (Single) & Cars (Multi) & Non-cars\\
    \toprule
    \textsc{Singlet} 
    & .311$\pm$.011 & .316$\pm$.011 & .324$\pm$.002 & .337$\pm$.002 & .335$\pm$.005 & .315$\pm$.002 \\
    \textsc{Singlet + BG}
    & .281$\pm$.010 & .277$\pm$.008 & .360$\pm$.003 & .360$\pm$.002 & .366$\pm$.005 & .341$\pm$.002 \\
    {\color{gray}\textsc{Best-of-singlets}}
    & {\color{gray}.304$\pm$.011} 
    & {\color{gray}.307$\pm$.011} 
    & {\color{gray}.314$\pm$.001} 
    & {\color{gray}.330$\pm$.002} 
    & {\color{gray}.324$\pm$.004} 
    & {\color{gray}.312$\pm$.004} \\
    {\color{gray}\textsc{Near. neigh.}}
    & {\color{gray}.277$\pm$.009} 
    & {\color{gray}.277$\pm$.009}
    & {\color{gray}.360$\pm$.002}
    & {\color{gray}.360$\pm$.002}
    & {\color{gray}.332$\pm$.007}
    & {\color{gray}.313$\pm$.004}\\
    \textsc{Mask-aware mean}
    & .290$\pm$.012 & .293$\pm$.012 & .306$\pm$.002 & .324$\pm$.002 & .305$\pm$.004 & .285$\pm$.002 \\
    \textsc{Triplets}
    & .268$\pm$.011 & .277$\pm$.011 & .313$\pm$.001 & .332$\pm$.002 & .284$\pm$.002 & .288$\pm$.001 \\
    \textsc{Triplets +  BG}
    & \textbf{.210$\pm$.007} & \textbf{.226$\pm$.007} & \textbf{.305$\pm$.001} &  \textbf{.315$\pm$.001} & \textbf{.272$\pm$.004} & \textbf{.279$\pm$.001} \\
    \bottomrule
	\end{tabular}
	\label{tbl:Results}
\end{table*}

\textbf{Numerical comparison}
All variants are run on all subsets of our test set: synthetic and real, both single and multi-material, for all objects. 
Results are summarized in \refTbl{Results}.
Note that, as the table columns refer to different cases, cross-column comparisons are generally not advised, but below we try to interpret the recovered results.
For the synthetic cars, we see a consistent improvement by adding background information already for the \textsc{Singlet} - even outperforming {\color{gray} \textsc{Best-Of-Singlets}}. 
Across all experiments, there is consistent improvement from \textsc{Singlet} to \textsc{Triplet} to \textsc{Triplet+BG}. 
\textsc{Triplet+BG} has consistently the best results - in particular outperforming the {\bf \color{gray} \textsc{Nearest Neighbor}}, which indicates generalization beyond the training set environment maps as well as the hand-crafted fusion scheme {\textsc{Mask-Aware Mean}}.
Overall, it is striking that performance for the multi-material case is very strong. 
This is appealing as it is closer to real scenarios. 
But it might also be counter-intuitive, as it seems to be the more challenging scenario involving multiple unknown materials with less observed orientations. 
In order to analyze this, we first observe that for \textsc{Singlet}, moving from the single to the multi-material scenario does not affect performance much. 
We conclude that our method is robust to such sparser observation of normals. 
More interestingly, our best performance in multi-material scenario is only partially explained by exploiting the ``easiest'' material, which we see from {\color{gray} \textsc{Best-Of-Singlets}}. 
The remaining margin to \textsc{Triplet} indicates that our model indeed exploits all 3 observations and that they contain complementary information.

\mycfigure{Alternatives_small}{Alternative approaches 
\emph{(left to right)}:
\textbf{1)} input(s).
\textbf{2, 4 and 6)} our approach for $n_\mathrm{mat}=1$.
\textbf{3, 5 and 7)} the same, including a background.
\textbf{8)} the nearest neighbor approach.
\textbf{9)} the mask-aware mean approach.
\textbf{10)} our approach for $n_\mathrm{mat}=3$.
\textbf{11)} the same, including a background, \ie full approach.
\textbf{12)} reference.
For a quantitative version of this figure see \refTbl{Results}.
For all images see supplementary material.
}

\textbf{Visual comparison}
Example outcomes of these experiments, are qualitatively shown in \refFig{Alternatives_small}.
Horizontally, we see that individual reflectance maps can indeed estimate illumination, but contradicting each other and somewhat far from the reference (columns labeled \textsc{Singlet} in \refFig{Alternatives_small}).
Adding the BG information can improve color sometimes (columns \textsc{+BG} in \refFig{Alternatives_small}).
We also see that a nearest neighbor oracle approach (column \textsc{NN} in \refFig{Alternatives_small}) does not perform well.
Proceeding with triplets (column \textsc{Triplet} in \refFig{Alternatives_small}) we get closer to the true solution. 
Further adding the background (\textsc{Ours} in \refFig{Alternatives_small}) results in the best prediction.
We see that as the difficulty increases from spheres over single- and multi-material to complex shapes, the quality decreases while a plausible illumination is produced in all cases.
Most importantly, the illumination can also be predicted from complex, non-car multi-material objects such as the dinosaur geometry as seen in the last row.
Supplementary material visualizes all the alternatives for the test dataset.

\begin{table}[t]
    \centering
    \caption{Reconstruction on different number of materials $n_\mathrm{mat}$.}
    \begin{tabular}{rrr}
    & Spheres & Cars (Single) \\
    \toprule
    \textsc{Singlet + BG}
    & .360$\pm$.003 & .360$\pm$.002 \\
    \textsc{Doublets + BG}
    & .320$\pm$.002 & .327$\pm$.002 \\
    \textsc{Triplets + BG}
    & .305$\pm$.001 & .315$\pm$.001 \\
    \textsc{Quadruplets + BG}
    & .309$\pm$.001 & .306$\pm$.001 \\
    \textsc{Quintuplets + BG}
    & \textbf{.292$\pm$.001} & \textbf{.295$\pm$.001} \\
    \bottomrule
    \end{tabular}
    \label{tbl:ResulstMaterialCount}
\end{table}

\textbf{Varying the number of materials}
In another line of experiments we look into variation of $n_\mathrm{mat}$ in \refTbl{ResulstMaterialCount}.
Here, the number of input reflectance maps increases from $1$ up to $5$.
In each case we include the background and run both on spheres and single-material cars, for which these data are available for $n_\mathrm{mat}>3$.
Specifically, we use the real singlets, that we combine into tuples of reflectance maps according to the protocol defined in \refSec{Dataset}. 
We see that, although we have not re-trained our network but rather copy the shared weights that were learned using $n_\mathrm{mat}=3$ materials, our architecture does not only retain efficiency across an increasing number of materials in both cases, but in fact uses the mutual information to produce even an increase in quality.
This is in agreement with observations that humans are better in factoring illumination, shape and reflectance from complex aggregates than for simple ones \cite{vangorp2007influence}.

\textbf{Further analysis}
To assess the magnitude of the recovered dynamic range in \refFig{AnalysisPlots_small}~a we plot the distribution of luminance over the test dataset (yellow color) and compare it to the distribution of estimated illuminations (red color).
Despite the fact that our method operates using only LDR inputs, we observe that in the lower range the graphs overlap (orange color), but in the higher we do not reproduce some brighter values found in the reference.
This indicates that our results are both favorable in structure as seen from \refTbl{Results} and \refTbl{ResulstMaterialCount} as well as according to more traditional measures such as log L1 or L2 norms \cite{Lombardi2012, Lombardi2015}.

We also evaluate the spatiality of the recovered illumination, \ie dominant light source direction. 
In 97.5\% of the test dataset environment maps, the estimated brightest pixel (dominant light) is less than 1 pixel away from ground-truth, which indicates a fairly accurate prediction.

\myfigure{AnalysisPlots_small}{
Result analysis:
\textbf{a)} Predicted vs.\ ground-truth luminance distribution,
\textbf{b)} Visual quality w.r.t.\ material specularity,
\textbf{c)} DSSIM error w.r.t.\ reflectance map sparsity.
}

From the numerical and visual analysis presented above and in supplementary material we can extract useful insights.
First, we plot the DSSIM w.r.t. the sparsity of the input reflectance maps on the test set (see Fig.~\ref{fig:AnalysisPlots_small}~c).
We observe that for sparser reflectance maps, the DSSIM becomes higher (the error increases).
For sparser reflectance maps the network has more unobserved normals (pixels in the reflectance map) to hallucinate, making inference relatively harder.
Second, we study the visual quality of the results w.r.t. material attributes like specularity.
Fig.~\ref{fig:AnalysisPlots_small}~b visualizes from left to right the recovered visual details for an increasing specularity.
The visual quality increases as the more specular a material is, the more it reveals about the environment.

\myfigure{comparison_nishino}{
Visual comparison with~\cite{Lombardi2012,Lombardi2015}. Images were taken from the respective papers. \cite{Lombardi2012,Lombardi2015} use HDR inputs, we use LDR.
}

\textbf{Comparison with related work}
It is important to explain why existing approaches are not directly applicable in our case.
\cite{Lombardi2012, Lombardi2015} do not handle multiple materials.
One might argue that they can still be applied to the segmented subregions, but then it is unclear how to merge the different generated outputs; the papers do not provide a solution.
In this case, one would still need techniques like {\color{gray} \textsc{Best-Of-Singlets}} or {\textsc{Mask-Aware Mean}} to proceed.
Instead, our method naturally fuses the features from the segmented materials and background producing a single output.
\cite{lombardi2012single} uses multiple materials but works for point light sources and can not be assumed to extend to natural illumination.
\cite{LombardiN16} works with multiple single-material objects under natural illumination but requires multiple images as input ($\geq$ 3 according to the authors).
Most importantly, all of \cite{lombardi2012single, Lombardi2012, Lombardi2015, LombardiN16} require HDR images as input (\ie taking multiple pictures under different exposures) making them unworkable for our single-shot LDR input, and do not leverage the image background that is naturally captured in their images anyway.

Nevertheless, we provide an indicative visual comparison between our method and \cite{Lombardi2012, Lombardi2015} for the dataset of \cite{Lombardi2012} in~\refFig{comparison_nishino}. 
While \cite{Lombardi2012, Lombardi2015} are able to capture the major light components of the scene, our approach recovers detailed structures not only from the lights but also from the surrounding elements (\eg buildings and trees). 

\mysubsection{Qualitative results and applications}{Qualitative}

\myfigure{RealResults_small}{
	Results for our approach on real objects: spheres, single-material cars, multi-material cars, multi-material non-cars.
	\textbf{a)} input LDR images.
	\textbf{b)} our predicted HDR environment.
    \textbf{c)} the ground-truth.
    Exhaustive results are found in supplementary material.
}

The visual quality is best assessed from \refFig{RealResults_small}, that shows, from left to right, the input(s) (a), our estimated (b) and ground-truth environment map (c). 
The difficulty ranges: starting from spheres, we proceed to scenes that combine three single-material objects over single objects with multiple materials to non-car shapes with multiple materials.
This shows how non-car shapes at test time can predict illumination, despite training on cars and car parts.
In each case, a reasonable estimate of illumination is generated as seen from the two last columns in \refFig{RealResults_small} and supplementary material.

\myfigure{Relighting_small}{
	Comparison of re-rendering using the reference, ours, and nearest neighbor for a specular material.
    Ours is more similar to the reference, while not requiring to acquire an HDR light probe.
}

\textbf{Relighting} 
To verify the effectiveness in a real relighting application, we show how re-rendering with a new material looks like when illumination is captured using our method vs. a light probe.
In the traditional setup (which we also used to acquire the reference for our test data) one encounters multiple exposures, (semi-automatic) image alignment, and a mirror ball with known reflectance and geometry.
Instead, we have an unknown object with unknown material and a single LDR image.
Note how similar the two rendered results are in \refFig{Relighting_small}. 
This is only possible when the HDR is also correctly acquired.
Instead, a nearest-neighbor oracle approach already performs worse; the reflection alone is plausible, but far from the reference.
For more relighting examples see the supplemental video.

\myfigure{Usecase_small}{
	Material/shape edits on web images (see the text below).
}

\textbf{Images from the web}
Ultimately, our method is to be used on images from the web.
\refFig{Usecase_small} shows examples of scenes (top row), on which we run our method and then re-render editing either the material (middle row) or the shape (bottom row).
The rendered results look nevertheless convincing.
The supplementary material contains more results that indicate the method's performance on everyday images.

\mysection{Conclusion}{Conclusion}
We have shown an approach to estimate natural illumination in HDR when observing a shape with multiple, unknown materials captured using an LDR sensor.
We phrase the problem as a mapping from reflectance maps to environment maps that can be learned by a suitable novel deep convolution-de-convolution architecture we propose.
Training and evaluation are both made feasible thanks to a new dataset combining both synthetic and acquired information. 
Due to its learning-based nature, we believe that our approach can be possibly extended to compensate for known material segmentation and geometry, and achieve higher fidelity results if trained on a larger "in the wild" dataset.

\textbf{Acknowledgements} This work was supported by the FWO (G086617N) and the DFG (CRC 1223).

{\footnotesize
\bibliographystyle{ieee}
\bibliography{paper}
}

\end{document}